\documentclass[conference]{IEEEtran}
\IEEEoverridecommandlockouts
\usepackage{booktabs} 
\usepackage[table,xcdraw]{xcolor}
\usepackage{multirow}      
\usepackage{multicol}
\usepackage{soul}
\usepackage{array}

\usepackage{verbatim} 
\newcolumntype{C}[1]{>{\centering\arraybackslash}m{#1}}  

\usepackage{cite}

\ifCLASSINFOpdf
  \usepackage[pdftex]{graphicx}
\else
\fi
\ifCLASSOPTIONcompsoc
 \usepackage[caption=false,font=normalsize,labelfont=sf,textfont=sf]{subfig}
\else
 \usepackage[caption=false,font=footnotesize]{subfig}
\fi
\usepackage{url}

\begin{document}
%
\title{Predicting Abandonment in Online Coding Tutorials}

\author{\IEEEauthorblockN{An Yan}
\IEEEauthorblockA{The Information School\\
University of Washington\\
Seattle, Washington, 98105\\
yanan15@uw.edu}
\and
\IEEEauthorblockN{Michael J. Lee}
\IEEEauthorblockA{Department of Informatics\\
New Jersey Institute of Technology\\
Newark, New Jersey, 07102\\
mjlee@njit.edu}
\and
\IEEEauthorblockN{Andrew J. Ko}
\IEEEauthorblockA{The Information School\\
University of Washington\\
Seattle, Washington, 98105\\
ajko@uw.edu}}

\IEEEpubid{978-1-5386-0443-4/17/\$31.00~\copyright~2017 IEEE}


%


\maketitle

\begin{abstract}
Learners regularly abandon online coding tutorials when they get bored or frustrated, but there are few techniques for anticipating this abandonment to intervene. In this paper, we examine the feasibility of predicting abandonment with machine-learned classifiers. Using interaction logs from an online programming game, we extracted a collection of features that are potentially related to learner abandonment and engagement, then developed classifiers for each level. Across the first five levels of the game, our classifiers successfully predicted 61\% to 76\% of learners who did not complete the next level, achieving an average AUC of 0.68. In these classifiers, features negatively associated with abandonment included account activation and help-seeking behaviors, whereas features positively associated with abandonment included features indicating difficulty and disengagement. These findings highlight the feasibility of providing timely intervention to learners likely to quit.\end{abstract}
\ifCLASSOPTIONpeerreview
\begin{center} \bfseries EDICS Category: 3-BBND \end{center}
\fi
%
\IEEEpeerreviewmaketitle

\section{Introduction} \label{intro}
The popularity of online coding tutorials such as Codecademy, Kahn Academy, and Code.org has risen dramatically in past years, reaching millions of people eager to learn to code. However, because these resources are discretionary and frequently lack the social and instructional support that classroom environments provide, learners often abandon them at high rates \cite{kizilcec_attrition_2015,lee_-game_2013}.

There are many ways to encourage learners to continue learning in these tutorials. Prior work has investigated improving error message feedback \cite{lee_personifying_2011}, incorporating assessments to validate learners' understanding \cite{lee_-game_2013}, and tuning material difficulty \cite{gutl_attrition_2014}. These efforts have proven effective at increasing how far a learner gets through online materials before quitting.

Unfortunately, some of the most effective strategies for increasing engagement are still limited to classroom environments. For example, one of the simplest and most powerful strategies is encouragement: when a learner is stuck, a few supportive words can promote further engagement, causing increases in self-efficacy and intrinsic motivation, especially to students who start with low self-efficacy for a skill \cite{hitz_praise_1988,stake_critical_2006,tuckman_effect_1991}. Encouragement can have positive effects on learning and engagement even when coming from software \cite{brown_positive_2014}, suggesting that it is the content and context of encouragement, and not who it comes from, that is important.

Offering encouragement in coding tutorials poses many challenges. If encouragement is not timed well, it may interrupt learners' thoughts, causing them to lose their places in a difficult coding problem or exercise. Moreover, learners who are already engaged and successfully learning may perceive encouragement as annoying, condescending, or as a lack of confidence in their abilities, making them question whether they really are succeeding. This leaves designers of online coding tutorials between two extremes: if they always provide encouragement, they may annoy the majority of learners who are engaged; if they never provide encouragement, they fail to help learners who are likely to abandon the tutorial. Naive probabilistic models also have flaws: for a coding tutorial with a 10\% abandonment rate on each lesson, we could randomly predict that 1 in 10 learners would quit, but this would only correctly identify 1\% of learners about to leave. This would also unnecessarily encourage many learners.
%
%
In this paper, we explore the feasibility of predicting which learners are likely to complete the next lesson of a tutorial, identify features of learner behavior that best inform this prediction, and examine whether these features show consistent positive or negative associations with abandonment. While these feasibility assessments do not yet directly explore the applications to encouragement in online tutorials, they provide a foundation for future efforts.
\IEEEpubidadjcol 
\section{Related Work} \label{sec:related}
\subsection{Dropout in Introductory Programming Courses}
It is well established that introductory programming (CS1) courses in higher education have high dropout rates \cite{guzdial_teaching_2002,kinnunen_why_2006}. A worldwide survey on pass rates reported that, on average, only 67\% of students complete their CS1 course \cite{bennedsen_failure_2007}. Watson \& Li expanded on this study, finding that the mean pass rate of CS1 courses is 67.7\% and that pass rates have not improved over time \cite{watson_failure_2014}.

Several studies have investigated the reasons for these pass rates. Kinnunen \& Malmi conducted interviews with 18 dropouts from a CS1 course at Helsinki University of Technology \cite{kinnunen_why_2006}, which showed that a lack of time and a lack of motivation were the key reasons for dropping out. These two factors are affected by perceived difficulty of the course and difficulties with time management. Other factors that influence dropout included gender, prior programming experience, students' attributions of success, learning style, mental model of programming, and self-efficacy \cite{ramalingam_self-efficacy_2004}.

Some studies have attempted to predict dropout rates in CS1 courses based on these factors. Wilson \& Shrock examined twelve factors of success in a CS1 course \cite{wilson_contributing_2001}. They surveyed 105 students at a Midwestern university and found that ``comfort level,'' math knowledge, and attribution to luck were the most predictive of success. Similarly, Ventura Jr. found that a math background, prior programming experience, and gender have negligible predictive power in the success of a CS1 course, whereas student effort and comfort were the strongest predictors of success \cite{jr_identifying_2005}.

Although researchers and educators have investigated student attrition in CS1 courses extensively, there is little consensus on which factors are most significant in predicting dropout. Moreover, many factors that appear to have significance in classroom contexts are more difficult to observe in online contexts, since they require measuring learners' social context, identity, and prior experience.
\subsection{Dropout in MOOCs} 
Massive open online courses (MOOCs) have increased in popularity over the last few years, particularly those teaching computing. Of the many challenges that MOOCs face, one is the low user retention rates. Studies have shown that approximately 5\% to 15\% of students registered for a course complete it~\cite{ramesh_modeling_2013,taylor_likely_2014}.

Studies of MOOC retention have many parallels to CS1 attrition rates. A survey of 134 MOOC users who dropped out showed that a majority of them reported changes in job, lack of time, content difficulty, and also a lack of difficulty as the reasons for leaving \cite{gutl_attrition_2014}. Kizilcec \& Halawa surveyed 550 respondents who were identified by a prediction model to be disengaged from a MOOC course. The results indicated non-behavioral characteristics such as gender, age, and geographical location were related to retention \cite{kizilcec_attrition_2015}.

Statistical models of MOOC dropout have also demonstrated some reasons why learners quit a course. Factors include student demographic characteristics, self-reported commitments, features of forum discussion activities, course performance, and students' usage patterns \cite{balakrishnan_predicting_2013,ramesh_modeling_2013}. Adamopoulos \cite{adamopoulos_what_2013} applied logistic regression to model course completion with data from 842 students enrolled in 133 courses offered by 30 universities. His results showed that student perceptions of the instructor, assignments, and course material were the most significant predictors of retention. Hermans \& Aivaloglou also applied logistic regression to predict completion of a MOOC course that teaches young children programming and software engineering concepts. The prediction results showed that factors related to course performance had positive influence on completion, whereas parental involvement and being late joining the class had negative influence on completion \cite{hermans2017teaching}. However, these two studies treated dropout as a single categorical variable, rather than examining the likelihood of dropout at different points over time. 

Using survival analysis to model retention in a Coursera discussion forum over the entire course period on a weekly basis, Yang et al. found that social interactions among students in a MOOC affected dropout throughout the duration of the course \cite{yang_turn_2013}. Instead of focusing on factors captured during the course, Greene et al. used survival analysis to predict dropout based on self-reported features gathered prior to the course, finding that age, prior experience, and self-reported commitment had significant predictive power \cite{greene_predictors_2015}.

A number of  studies have used machine learning to predict MOOC dropout. Xing et al. used ensemble learning based on General Bayesian Network and Decision Tree algorithms with features generated from a MOOC discussion forum, achieving accurate identification of at-risk students; it did not examine which features best predicted abandonment \cite{xing_temporal_2016}. Halawa et al. designed a prediction model using behavioral data with MOOC materials, finding that students' lack of interest and lack of ability accounted for 60\% of high-risk students two weeks before dropout \cite{halawa_dropout_2014}. Taylor et al. carried out a study of 70 gigabytes of student usage logs including click-streams, forum posts, wiki revisions, and learner state information from a MOOC course \cite{taylor_likely_2014}. They extracted 25 predictive features using crowd-sourcing methods, and derived over 10 thousand models with various machine learning techniques on a per student and per week basis. Their model achieved a notably high AUC (area-under-the-curve of the receiver operating characteristic, a metric to evaluate a binary classifier system) of 0.95. Features related to homework submission, social interactions, and lab grades were the most predictive. 

%

%

\subsection{Abandonment in Coding Tutorials}
There has been less work investigating dropout in online coding tutorials. Abandonment of tutorials is a slightly different problem, as learners who use tutorials are not necessarily making the same commitment as when enrolling in a course. Therefore, rather than using the word ``dropout,'' we prefer using the word ``abandon,'' which does not assume that someone using a tutorial has committed to learn.

One of the most recent works on engagement in online coding tutorials is that by Lee et al. on the Gidget programming game \cite{lee_comparing_2015-1}. Lee investigated several strategies for preventing abandonment, including more personal errors \cite{lee_personifying_2011} and the inclusion of in-game assessments \cite{lee_-game_2013} -- both significantly increase engagement. Repenning et al. used their \textit{Retention of Flow} framework to analyze Hour of Code activity data collected from 5,512 student projects during CSEdWeek 2014, where students viewed a tutorial and built a Frogger game using an online programming environment. Their analysis showed that the loss of retention might result from cognitive, technical, and practical challenges \cite{repenning2016retention}. Outside programming, researchers have built predictive models of learners' motivational states in similar interactive learning environments \cite{cocea_eliciting_2007,qu_detecting_2005}, finding features related to help-seeking with manuals and tool-tips to have strong predictive power.
    
These and other efforts from prior work have several implications for coding tutorial abandonment prediction. First, many of the most predictive features in prior work have concerned social, instructional, and motivational factors, all of which are difficult to detect in coding tutorials, especially if used anonymously. Moreover, the majority of studies have considered dropout at the \textit{end} of a course in learning, leaving open the possibility that early detection of dropout is not feasible. That said, prior work suggests that some behavioral features, such as different types of learners' actions, may be strong indicators of either engagement or disengagement. In this paper, we will assess the feasibility of predicting abandonment in an online coding tutorial, investigate which features, if any, are most predictive of abandonment, and how they influence a learner's decision to abandon the tutorial. 

\section{Method}
In this section, we detail the online tutorial for which we predicted abandonment, introduce the dataset, and describe our classification approach.
\begin{figure}
\centering
\includegraphics[width=0.45\textwidth]{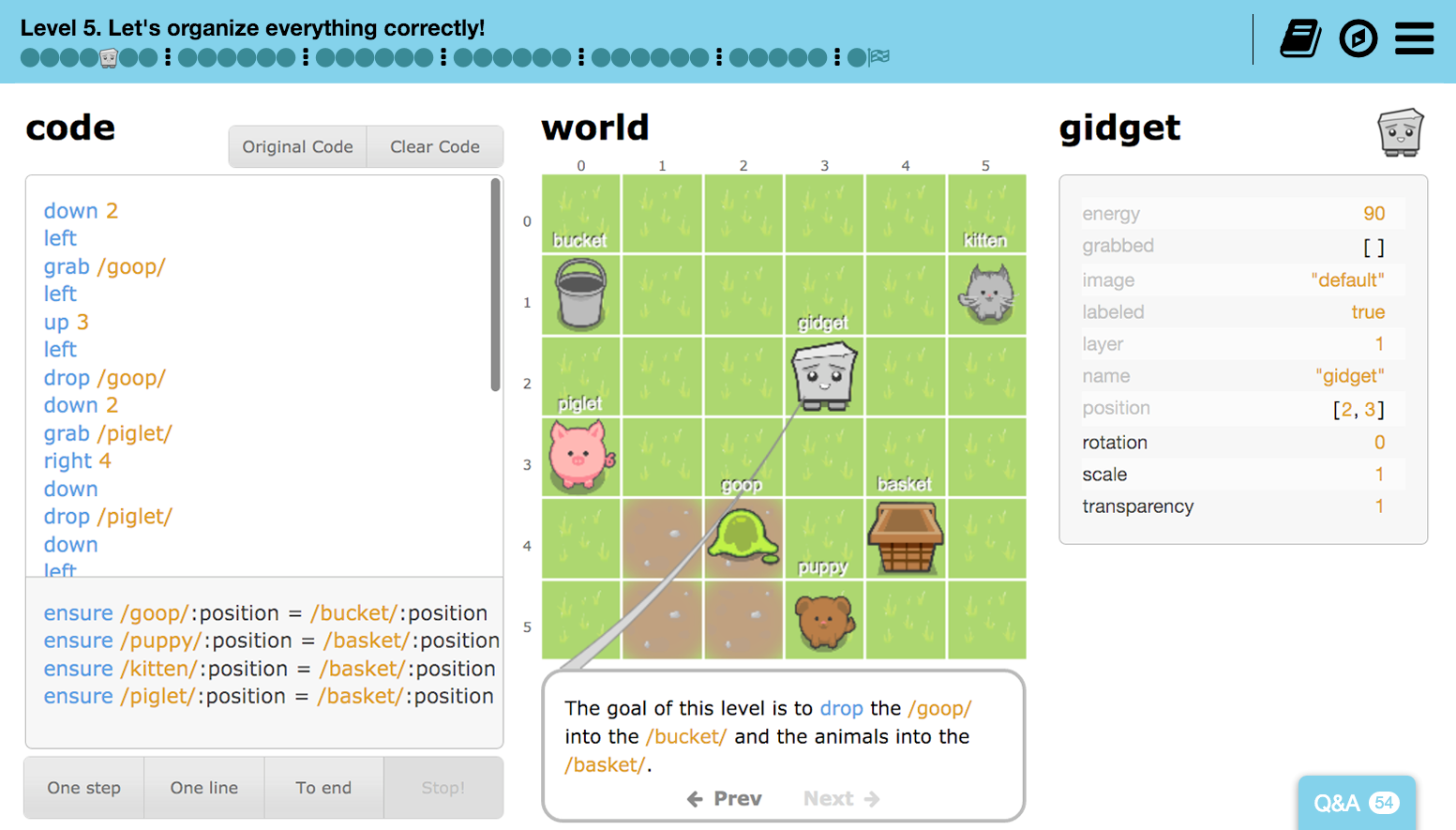}
 \caption{Gidget's Level 5, showing the code editor, level goals, objects in the world container, mission text, and runtime state. Gidget logs all interactions with all components.}\label{fig:gidget-interface}
\end{figure}
\begin{figure}[b]
\centering
\includegraphics[width=0.45\textwidth]{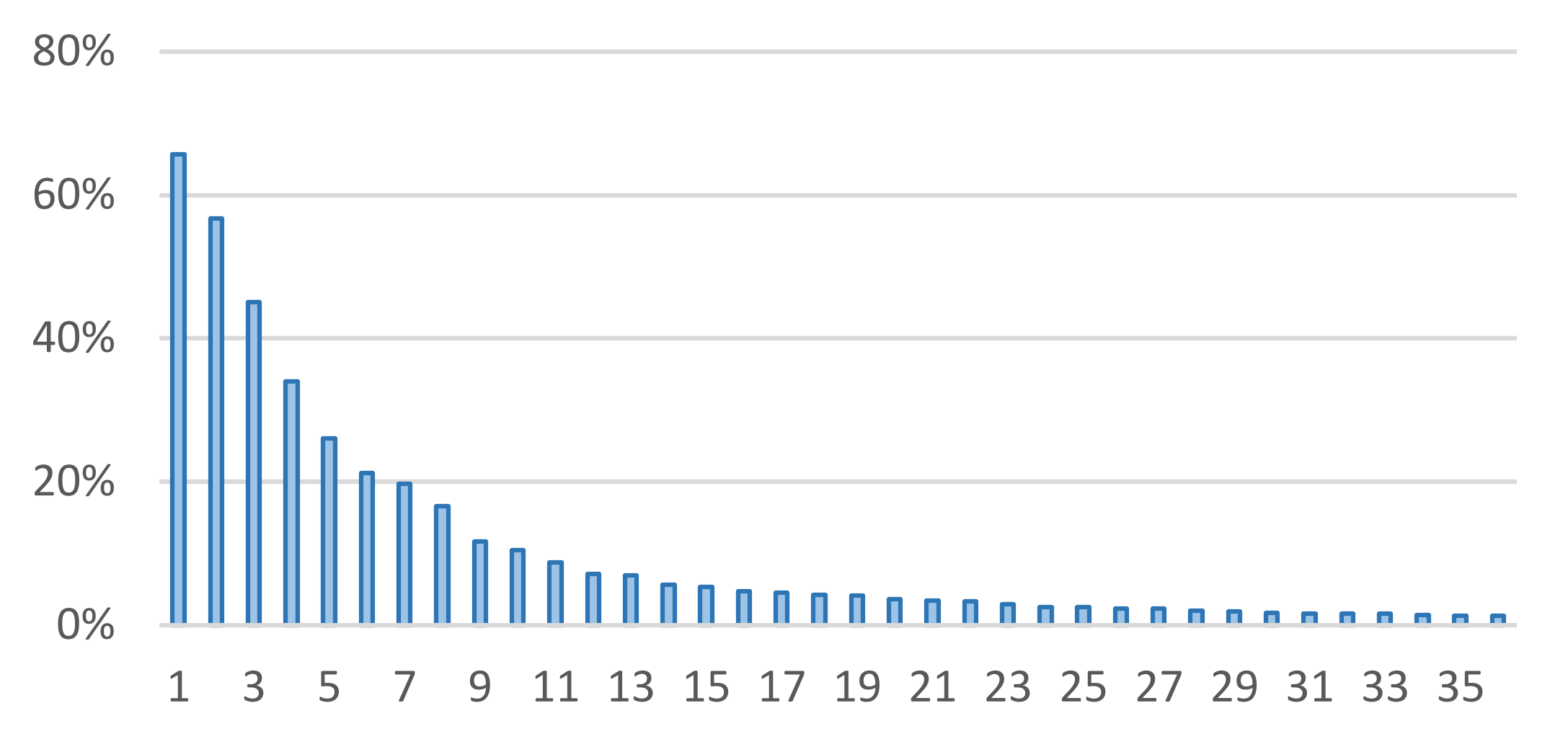}
 \caption{Percent of learners remaining after each level (N = 5,038).}\label{fig:stay-rate}
\end{figure}
\subsection{The Gidget Programming Game}

\begin{figure}[b]
\centering
 \includegraphics[width=0.45\textwidth]{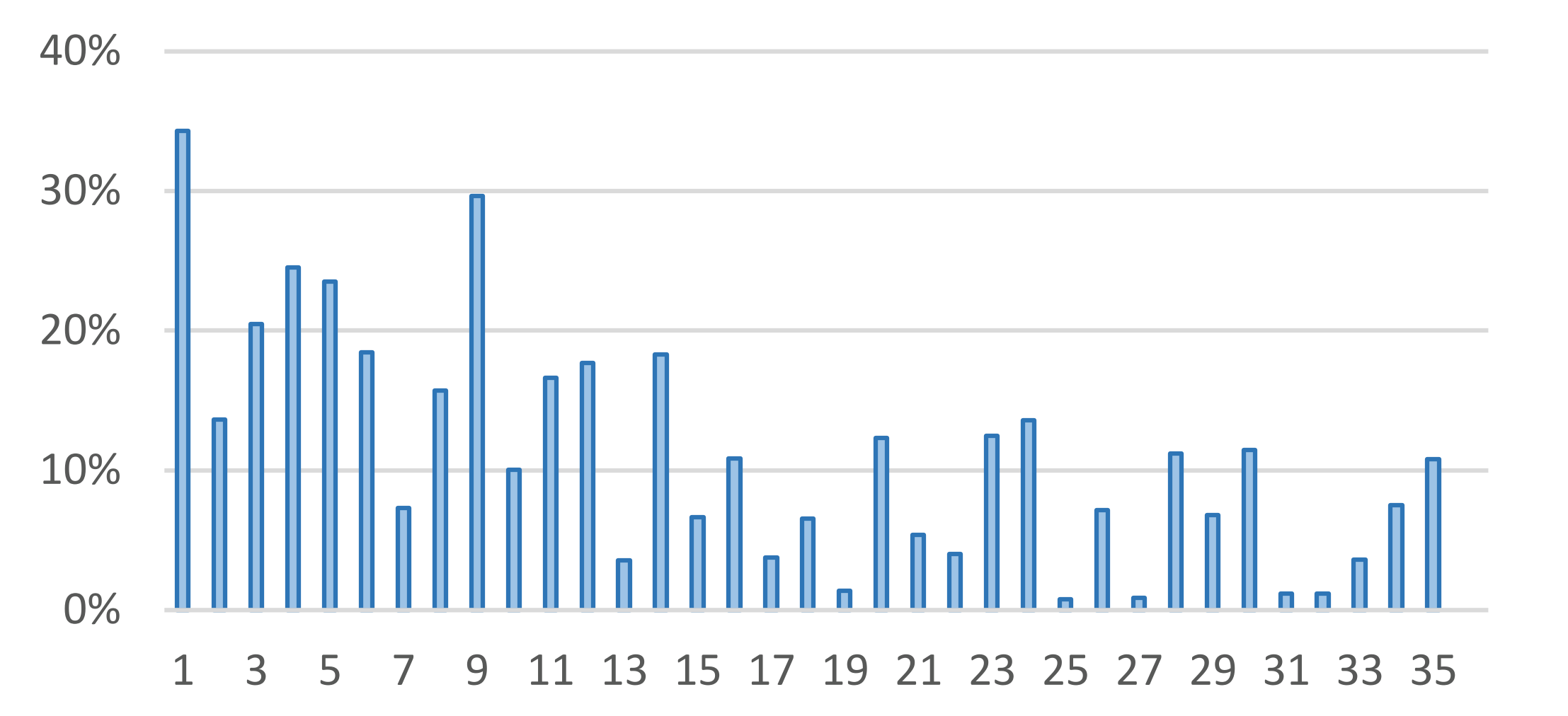}
 \caption{Learners' abandonment rate on each level, ranging from 0-34.3\%.}\label{fig:abandon-rate}
\end{figure}

\begin{table*}
  \caption{Features extracted for abandonment prediction.}
  \label{tab:features}
  \centering
\begin{tabular}{@{}p{2.2cm}lp{7.8cm}l@{}}
\toprule
 & FEATURE & DEFINITION & RATIONALE \\ \midrule
\multirow{11}{*}{\begin{tabular}[c]{@{}p{2.2cm}@{}}\textbf{CUMULATIVE FEATURES}\\   Based on activities in the current and prior levels.\end{tabular}} & cml\_total\_dur & \begin{tabular}[c]{@{}p{8cm}@{}}Cumulative total non-idle time spent on all activities\end{tabular} & Indicates effort \\ 
 & cml\_idle\_time & \begin{tabular}[c]{@{}p{7.8cm}@{}}Cumulative total idle-time lacking mouse and keyboard events\end{tabular} & Suggests disengagement \\  
 & cml\_code\_time & \begin{tabular}[c]{@{}p{7.8cm}@{}}Cumulative total time code editor had  keyboard focus\end{tabular} & Suggests persistence \\ 
 & cml\_test\_time & \begin{tabular}[c]{@{}p{7.8cm}@{}}Cumulative total time spent in program execution mode\end{tabular} & Suggests desire to succeed \\  
 & cml\_help\_time & \begin{tabular}[c]{@{}p{7.8cm}@{}}Cumulative total time focused on tutorial and reference guide\end{tabular} & Suggests desire to learn \\ 
 & cml\_mission\_time & \begin{tabular}[c]{@{}p{7.8cm}@{}}Cumulative total time spent on reading  mission texts\end{tabular} & Suggests engagement \\ 
 & \begin{tabular}[c]{@{}l@{}}cml\_world\_time\end{tabular} & \begin{tabular}[c]{@{}p{7.5cm}@{}}Cumulative total time spent on world container\end{tabular} & Suggests engagement \\ 
 & cml\_n\_restart & \begin{tabular}[c]{@{}p{7.5cm}@{}}Cumulative sum of \# of stop button and retry button clicks\end{tabular} & Indicates difficulty \\ 
 & cml\_n\_step & \begin{tabular}[c]{@{}p{7.5cm}@{}}Cumulative \# of one-step execution button  clicks\end{tabular} & Indicates difficulty \\ 
 & cml\_n\_line & \begin{tabular}[c]{@{}p{7.5cm}@{}}Cumulative \# of one-line execution button clicks\end{tabular} & Indicates difficulty \\ 
 & cml\_n\_play & Cumulative \# of ``run to end'' button clicks & Indicates evaluation activities \\ \midrule
\multirow{4}{*}{\begin{tabular}[c]{@{}p{2.2cm}@{}}\textbf{LEARNER} \\ \textbf{FEATURES}\\ Gathered upon registration.\end{tabular}} & activated & \begin{tabular}[c]{@{}p{7.8cm}@{}}Player registered an account (1), or did not (0)\end{tabular} & Indicates commitment to learn \\ 
 & \st{experience} & Prior experience programming (1), or none (0) & Predictive in {\cite{greene_predictors_2015,kizilcec_attrition_2015}} \\ 
 & \st{age} & Age in years & Predictive in {\cite{greene_predictors_2015,kizilcec_attrition_2015}} \\ 
 & \st{gender} & Male or female & Predictive in {\cite{greene_predictors_2015,kizilcec_attrition_2015}} \\ \bottomrule
\end{tabular}
\end{table*}
We selected the Gidget programming game (shown in Figure \ref{fig:gidget-interface} and available at \textit{www.helpgidget.org}) as our tutorial for investigation, as the game was heavily instrumented to record learner activity. Gidget teaches CS1 programming concepts such as variables, conditionals, loops, and functions to novice learners, along with concepts of testing and debugging. The game is structured as a sequence of levels, each presenting a defective program for the learner to fix. Each level focuses on a specific programming language concept, providing instruction by encouraging learners to step through the program's execution (as when using an interactive debugger), while Gidget the Robot explains the execution of each step in the program. The game also includes assessments between units of the curriculum to test and reinforce understanding of the concepts taught in the unit \cite{lee_-game_2013}.

As with most coding tutorials, the goal of Gidget is to provide a basic understanding of a programming language. However, because it is a game, it does differ from other tutorials in a few notable ways. First, the level of detail given about program execution is much higher than in most interactive tutorials, which tend to provide basic error feedback but do not provide granular feedback. Unlike most tutorials, the game also provides a story, which aims to motivate learners to continue through the game \cite{6344507}. The environment also offers a syntax guide, abundant example code, and contextual hints, as described in prior work \cite{lee2014principles ,jernigan_principled_2015,jernigan2017general}.

Despite these engagement-oriented features, learners who find the game online still abandon it at high rates. Figure~2 shows the percent of learners remaining after each level of the game, showing that only a few percent of learners complete the entire game. Figure 3 shows the per-level abandonment rate, ranging from 0\% to 34.3\%, with a mean rate of 13.4\% per level. Levels that introduce new programming concepts generally have higher abandonment rates. Levels with lower abandonment rates include Gidget's assessment levels, which present multiple choice and open-ended questions along with answers that attempt to correct learners' misconceptions with the concepts presented earlier \cite{lee_-game_2013}.

Due to the minimum requirements required by our analyses and low number of data points for later levels, this paper focuses on predicting learner abandonment between Levels 1 through 5, which have the most learner activity. In these levels, the game taught commands for moving Gidget the Robot around the world and commands for picking up and dropping objects (which were necessary for achieving game goals like those shown in the bottom-left portion of Figure 1, requiring objects to be moved and placed on other objects). Level 6 assessed learners' knowledge of the prior levels, asking them to predict output of a program that used commands introduced in the first five levels. Subsequent levels introduced more advanced concepts, including conditionals, variables, loops, functions, and objects.

\subsection{Data Collection}
The Gidget game records learner activity, including discrete information for each learner, for each level, including data such as the total time spent on a level, the total number of times specific buttons were clicked, the total time help was used, the total time spent on different interface elements, all code edits, whether or not a level was completed, and so on. Creating an account is optional, but ensures the player can return to his or her game at a later time. When creating an account, learners indicate their e-mail address, age, gender, and whether or not they have prior programming experience. Those who choose not to create an account can still play through the entirety of the game, but would lose access to their progress after their web session expires, or their web browsers' cache is cleared.

We obtained a database snapshot of 5,038 unique learners' activity logs spanning 23,647 level plays, representing 20 months of activities of all learners. We extracted most of our features from these logs. Gidget's developers primarily relied on word-of-mouth to recruit learners. Internet Protocol (IP) address logs and Google Analytics data indicate that learners accessed the game from a total of 103 countries, with the large majority of visitors coming from the USA (65.2\%) and Russia (16.4\%). Aggregate Google Analytics data indicates that Gidget users consisted of 55.7\% males and 44.3\% females, which is consistent with previously reported user demographics \cite{lee2015teaching}. Google Analytics only collects data for users who are at least 18 years old, and reported that 22.7\% of Gidget's users were 18-24 years old, 38.6\% were 25-24 years old, 21.2\% were 35-55 years old, 10.2\% were 45-54 years old, and 7.1\% were 55+ years old. We exclude further demographic data in this study because most users did not provide it, preventing us from tying particular demographic attributes to individual users.
\subsection{Prediction Problem Definition}
We define abandonment as a binary outcome for each level: either a learner completed a level or they did not, as specified by the game's level completion rules (which consisted of one or more test case assertions about program state, as shown in the left portion of Figure \ref{fig:gidget-interface}). For example, learners who abandoned at Level 1 means they did not satisfy Level 1's tests, and learners who abandoned at Level 2 means that they finished Level 1 and began playing Level 2, but did not satisfy Level 2's tests. Learners had the option to replay previously completed levels, but we always use their highest level played when defining abandonment.

Our prediction problem was as follows: for a given completed Level \textit{n} (where \textit{n} $\geq$ 1), predict whether a learner who has completed that level will complete Level \textit{n}+1 based on features of the learner and the learner's activities in Levels 1 through \textit{n}. This prediction would allow the game to provide encouragement just after the completion of a Level \textit{n}, or at the beginning of Level \textit{n}+1.

\subsection{Feature Extraction}
To determine features for predicting abandonment, we considered five sources: 1) factors that appeared related to engagement from previous work on Gidget \cite{lee_personifying_2011,6344507,lee_-game_2013,lee_comparing_2015-1}; 2) factors from studies of MOOCs and in-person CS1 courses; 3) features that we brainstormed based on the available data in the Gidget activity logs; 4) features proven to be significant in existing work on MOOC dropout prediction; and 5) features that are straightforward to interpret and informative to game designers. Table \ref{tab:features} lists the 12 features from this brainstorming process. Features with strikethroughs were excluded because they had too many missing values.

For each of these features, we computed the cumulative values from activity logs for each level (excluding the \textit{activated} feature, which only had one value per learner). The cumulative features were the sum of per-level features in completed levels. For example, \textit{cml\_total\_dur} in Level 3 was the sum of \textit{total\_dur} in Level 1, Level 2, and Level 3.

The 12 features largely reflect factors reported in the prior discussed in Section \ref{sec:related}. For example, \textit{cml\_n\_step} is the cumulative number of time executing the program step-by-step, which measures a testing behavior. A large value of \textit{cml\_n\_step} might be a sign of difficulty. However, some of the features have ambiguous indications. For instance, \textit{cml\_test\_time}, which measures the total time a player spent on code execution, can either suggest difficulty or desire to succeed. We therefore lack an understanding of how important the features selected are in a learner's decision to stay and if they possess consistent signals of abandonment (either abandon or not) across levels. We answer this question by assessing feature importance and feature impact in Sections \ref{sec:importance} and \ref{sec:impacts}.

\subsection{Data Preprocessing}

Our data cleaning focused on excluding invalid values. We found that because most learners did not register an account, most learner features had large amounts of missing data: \textit{age} was missing 76.0\%, \textit{experience} was missing 86.7\%, and \textit{gender} was missing 77.4\%. We removed these features as there was no way to estimate their missing values. Then, we computed cumulative features from corresponding per level features. Next, for each level, we extracted features for only the set of learners who had completed the level, ensuring that predictions were only based on learners that were active on that level. This resulted in a smaller set of labeled data for each level, mirroring abandonment. Defects in logging had also caused some missing values in \textit{total\_dur}, \textit{n\_play},\textit{ n\_line}, \textit{n\_step}, \textit{n\_restart} and their corresponding cumulative features, but we retained them because they were missing less than 5\% of their values in each level.

Since we did not have a large dataset, we wanted to retain as much of the useful signal in the data without weakening the classifier performance or inducing bias. Therefore, we used K nearest neighbor (KNN) imputation, which has proven to be a simple but robust way of offering accurate missing value estimation \cite{troyanskaya_missing_2001}. We implemented KNN imputation with the R package ``VIM'' for each level's feature data.
Lastly, we normalized the imputed datasets as z-scores, giving all features a zero-mean and unit standard deviation. Normalization rescales features that have a wide range of values, preparing them for use by many machine learning algorithms that assume features with consistent scales. After preprocessing, we had 3,292 users for level 1 predictions, and 2,841, 2,256, 1,702, and 1,298 users for levels 2, 3, 4 and 5, respectively.

\subsection{Classifier Building and Evaluation} \label{sec:classifier}

After feature extraction, we built classifiers for Levels 1 to 5 using three classifiers: Logistic Regression with L1 regularization (LR), Random Forest (RF) from the Python Sklearn Library \cite{pedregosa_scikit-learn:_2011}, and Gradient Boosting Decision Tree (GBDT) algorithm implemented in the XGBoost system \cite{chen_xgboost:_2016,friedman_greedy_2001}. These classifiers are widely utilized machine learning techniques across numerous disciplines including education research, and have been used effectively for dropout prediction in many previous works  \cite{liang_machine_2016,taylor_likely_2014,thammasiri_critical_2014,huang_telco_2015}. Moreover, XGBoost has dominated the winning solutions in recent machine learning and data science challenges such as in the KDDCup and Kaggle challenges.

Logistic Regression is a log-linear model for binary classification based on a set of features. L1 regularization is a technique to avoid over-fitting and increase model generalizability. Random Forests~\cite{ho_random_1995} is an ensemble learning method for classification and regression based on decision trees. A decision tree can model a binary decision making process and can learn a higher order of interactions between features than Logistic Regression. Random Forests are constructed with decision trees by randomly sampling with replacements from training data (see Section \ref{sec:importance} for further explanation). Gradient Boosting Decision Tree is similar to Random Forest in that they are both a set of decision trees. While each tree in Random Forests is trained independently, GBDT adds one tree at a time aimed to minimize the errors from the already-trained set of trees. 

Each classifier predicted whether a learner would abandon the next level of the game based on cumulative features from prior levels plus learner features. For each level, we randomly split the dataset into a training set (80\%) and a test set (20\%). For the training set of each level, we applied 10-fold cross-validation, where 90\% of the data were used for training and 10\% reserved for evaluation in each of the ten total trials. 

We compared our classifiers with three baseline classifiers. One classifier (baseline 1) randomly predicted abandonment by respecting the label distribution of learners in training data at each level, a second classifier (baseline 2) always predicted abandonment, and a third classifier (baseline 3) never predicted abandonment. These baseline classifiers mirror the currently available alternatives to designers of online coding tutorials wanting to predict abandonment.
%
%

In evaluating the classifiers, we defined success as balancing both precision and recall (true positives, TP), weighing equally the encouragement of learners likely to abandon, while avoiding unnecessarily encouraging learners who were likely to complete the next level. Therefore, we computed a number of evaluation metrics, including precision, recall, and F1 (the harmonic mean of precision and recall) on a test dataset for each level. For all of these metrics, we treated a learner abandoning a level as ``positive'' and learner completing a level as ``negative.'' We also computed AUC to measure the robustness of the classifiers across different baseline rates of per-level abandonment. AUC curves represent the trade-off between the true positives and false positives and are independent of abandonment rate, therefore providing a single scalar metric to model comparison.

\begin{table*}
  \caption{GBDT performance metrics on Levels 1 through 5 predictions.}
  \label{tab:performance1}
  \centering
\begin{tabular}{@{}p{0.58cm}C{0.68cm}C{0.37cm}C{0.37cm}C{0.37cm}C{0.68cm}C{0.37cm}C{0.37cm}C{0.37cm}C{0.68cm}C{0.37cm}C{0.37cm}C{0.37cm}C{0.68cm}C{0.37cm}C{0.37cm}C{0.37cm}C{0.68cm}C{0.37cm}C{0.37cm}l@{}}
\toprule
 & \multicolumn{4}{c}{Level 1} & \multicolumn{4}{c}{Level 2} & \multicolumn{4}{c}{Level 3}& \multicolumn{4}{c}{Level 4}& \multicolumn{4}{c}{Level 5} \\ \midrule
 & \textbf{GBDT} & bl1 & bl2 & bl3 & \textbf{GBDT} & bl1 & bl2 & bl3 & \textbf{GBDT} & bl1 & bl2 & bl3& \textbf{GBDT} & bl1 & bl2 & bl3 & \textbf{GBDT} & bl1 & bl2 & bl3  \\ \midrule
AUC & \textbf{0.64} & 0.51 & 0.50 & 0.50 & \textbf{0.68} & 0.51 & 0.50 & 0.50 & \textbf{0.70} & 0.50 & 0.50 & 0.50 & \textbf{0.61} & 0.47     & 0.50     & 0.50     & \textbf{0.77} & 0.53     & 0.50     & 0.50  \\
Prec &\textbf{0.19} & 0.16 & 0.14 & / & \textbf{0.29} & 0.22 & 0.21 & / & \textbf{0.42} & 0.29 & 0.29 & /& \textbf{0.37} & 0.21     & 0.26     & /         & \textbf{0.37} & 0.23     & 0.18     & /  \\
Recall & \textbf{0.75} & 0.17 & 1.00 & 0.00 & \textbf{0.76} & 0.21 & 1.00 & 0.00 & \textbf{0.65} & 0.23 & 1.00 & 0.00 & \textbf{0.61} & 0.18     & 1.00     & 0.00     & \textbf{0.72} & 0.21     & 1.00     & 0.00   \\
\begin{tabular}[c]{@{}l@{}}F1\end{tabular} & \textbf{0.30} & 0.16 & 0.25 & / & \textbf{0.42} & 0.22 & 0.34 & / & \textbf{0.51} & 0.25 & 0.45 & /  & \textbf{0.46} & 0.20     & 0.42     & /         & \textbf{0.50} & 0.22     & 0.31     & /   \\ 
\bottomrule
\end{tabular}
\end{table*}


\subsection{Feature Importance} \label{sec:method-importance}
We evaluated feature importance by examining the contribution of a feature in constructing gradient boosted trees in GBDT. In a single decision tree, the feature importance is concerned with the improvement in accuracy brought by adding a feature as a split. GBDT are constructed by adding one new tree at a time based on what the classifier already learned with existing trees, and then the relative rank of a feature in GBDT can be assessed by summing accuracy reduction for this feature over all trees \cite{he_practical_2014}. The tree structure can reveal complex interactions among features because every splitting node should respect the condition made by its parent node and a feature can appear multiple times in a tree. 

However, the ranking only provides a rough picture of feature importance as there are several correlated features in our choice of input. Specifically, there is up to 68.8\% pair-wise correlation among \textit{cml\_n\_line}, \textit{cml\_n\_restart}, \textit{cml\_n\_play}, and \textit{cml\_n\_step} within all five levels. We did not discard any of these features because we did not know which would be more predictive than the others. Although the performance of GBDT is robust to correlated features, when it comes to ranking feature importance, it will try to attach a higher score to only one of them and lower scores to the other correlated ones\footnote{Understand your dataset with XGBoost (XGBoost 0.6 documentation): $http://xgboost.readthedocs.io/en/latest/R-package/discoverYourData.html\#feature-importance$}. 

\subsection{Feature Impact} \label{sec:method-impacts}
We also examined the direction of these relationships with abandonment: on some levels, they may have been positive and on some they may have been negative. To get a sense of how a feature impacted abandonment prediction, we ran Logistic Regression on training data of every level. We then used Odds Ratios (OR) of each feature to assess the direction of its impact \cite{szumilas_explaining_2010}. The goal of using Logistic Regression here is different from Logistic Regression in Section \ref{sec:classifier}, the former is to interpret features, while the latter is to achieve good performance in classification tasks.

\section{Results}
In this section, we evaluate the classifiers, discuss the relative importance of their features in prediction, and assess how these features were associated with abandonment.
\subsection{Classifier Performance}
We report here in detail our best results from GBDT using cumulative features and learner features (12 features in total). Table~\ref{tab:performance1} shows the results for the five per-level classifiers, along with the results for the three baseline classifiers, indicated as ``bl1,'' ``bl2,'' and ``bl3'' for each level.
%
%

The AUC scores (Table \ref{tab:performance1}, row 1) across the Level 1 to 5 classifiers ranged from 0.61 to 0.77 with a mean of 0.68. A random binary predictor has an AUC of 0.5, which is usually used as a baseline for prediction evaluation. Therefore, our classifiers did better than chance at predicting abandonment. That said, maximum AUC is not always the optimal choice for a problem~\cite{lobo_auc:_2008}.

Our precision results (Table \ref{tab:performance1}, row 2) represent the percent of all learners our classifiers predicted would abandon the next level that actually did. From an encouragement perspective, precision represents the proportion of learners that would receive encouragement that actually needed it.  Our classifiers' precisions ranged from 0.19 to 0.42. The low precision for Level 1 is not surprising since there was very little on which to predict abandonment and the data was skewed with very high abandonment to non-abandonment ratios (1:6 for Level 2). This low precision means that the game might unnecessarily encourage many learners. Similarly, false positive (FP) rates indicate the percent of learners that completed the next level that we incorrectly predicted would abandon. This would be the percent of learners receiving encouragement that did not need it.  We attained an average FP of 0.37 across all 5 levels of prediction, suggesting that we might unnecessarily encourage 37\% of learners.
%
%

Table \ref{tab:performance1} (row 3) shows the recall of our classifiers. In the context of abandonment prediction, recall was the percent of all learners that abandoned a level that our classifiers successfully identified. From an encouragement perspective, this would be the percent of learners who needed encouragement that our classifiers successfully encouraged. Over all levels, the recall of our classifiers was above 0.6 with an average of 0.70, showing that the classifiers could identify on average about 70\% of learners likely to abandon. However, within the classifiers' limited prediction capability, there is always a tradeoff between TP and FP, meaning the higher percent of learners we successfully encouraged, the higher percent of learners we unnecessarily encouraged, and vice versa.
%
%

\begin{table}[b]
  \caption{Performance on Levels 1--5 predictions of Logistic Regression and Random Forests.}
  \label{tab:performance-mix}
  \centering
\begin{tabular}{@{}p{0.45cm}C{0.39cm}C{0.39cm}C{0.39cm}C{0.39cm}C{0.39cm}C{0.39cm}C{0.39cm}C{0.39cm}C{0.39cm}l@{}}
\toprule
    & \multicolumn{2}{c}{Level 1} & \multicolumn{2}{c}{Level 2} & \multicolumn{2}{c}{Level 3} & \multicolumn{2}{c}{Level 4} & \multicolumn{2}{c}{Level 5} \\ \midrule
    & LR           & RF          & LR           & RF           & LR           & RF           & LR           & RF           & LR           & RF           \\ \midrule
AUC & 0.61         & 0.62        & 0.64         & 0.64         & 0.65         & 0.64         & 0.60         & 0.61         & 0.65         & 0.66         \\
F1  & 0.30         & 0.32        & 0.42         & 0.42         & 0.53         & 0.52         & 0.45         & 0.46         & 0.40         & 0.43        \\ \bottomrule
\end{tabular}
\end{table}

The F1 scores reported in row four of Table \ref{tab:performance1} measure the balance between precision and recall, or, the balance between necessary and unnecessary encouragement. Our classifiers produced varied F1 scores ranging from 0.30 to 0.51, with our lowest performing classifiers again on Level 1 and 2 because of the low precision on early levels. This suggests that while achieving a balance between these goals is possible on many levels of the game, the classifiers did not perform well until they had enough data from prior levels to have a strong signal of engagement. While these classifier results are far from optimal, they are consistently better than the three baseline classifiers according to AUC scores and F1 scores.

Table \ref{tab:performance-mix} shows the results from Logistic Regression with L1 penalty and Random Forests for each of the five levels. Although inferior to GBDT, these classifiers outperformed the three baselines, demonstrating the set of features chosen influence prediction. Additionally, we ran GBDT with a set of 23 features consisting of 11 cumulative features, 11 per-level features, and 1 learner feature across Levels 1 through 5. The extra per-level features did not lead to a noticeable increase or decrease in the classifiers' performance.
\subsection{Feature Importance} \label{sec:importance}
\begin{figure}
\centering
\includegraphics[width=0.4\textwidth]{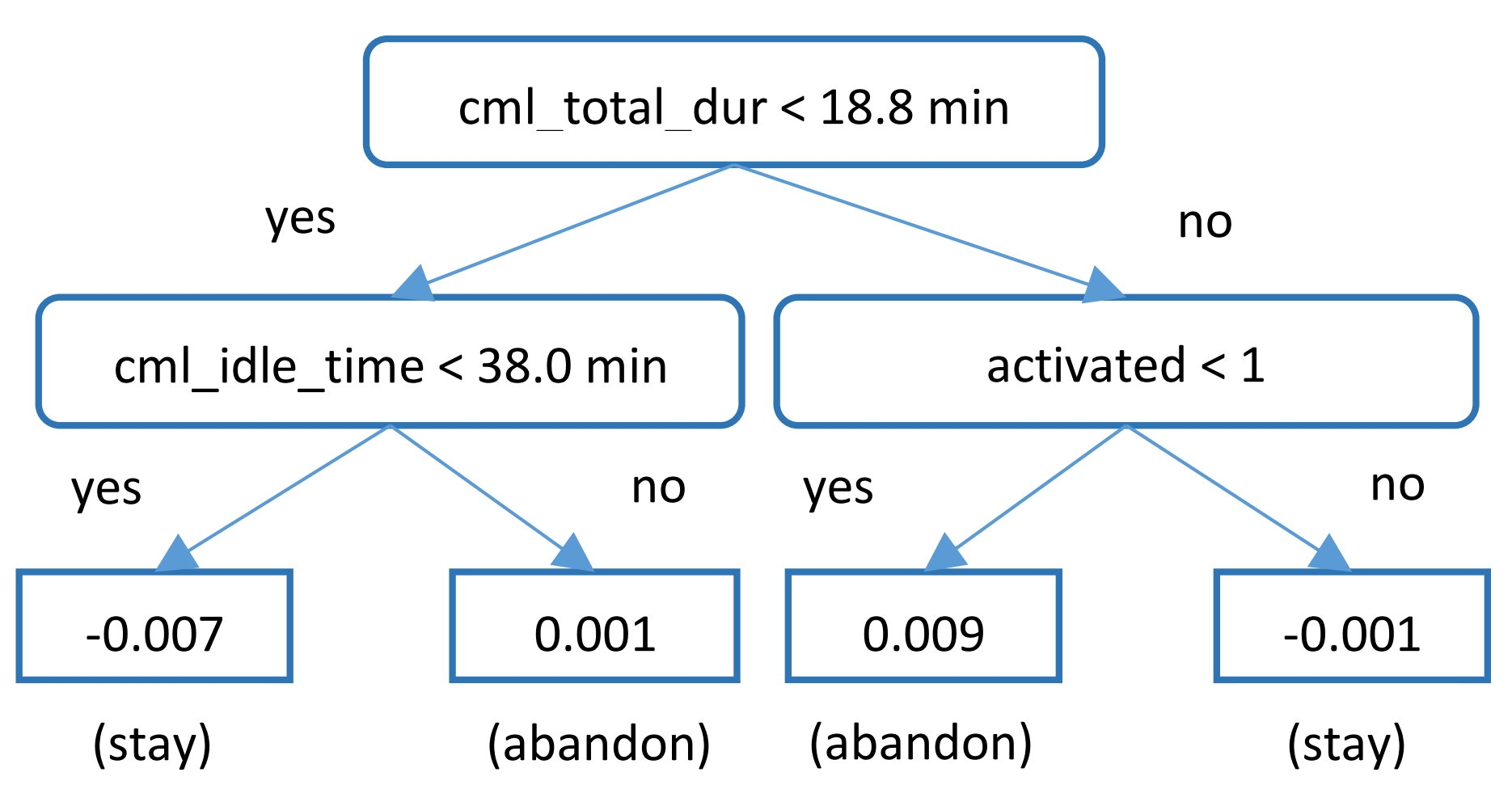}
 \caption{A decision tree in GBDT for Level 5 prediction.}\label{fig:tree}
\end{figure}
Having demonstrated the feasibility of predicting abandonment, we now turn to the underlying features that were most predictive of abandonment. First, we illustrate with an example how GBDT works in classification and determining feature importance. Figure \ref{fig:tree} shows one of the many decision trees with depth = 2 built by GBDT for a Level 5 prediction. Each leaf node is associated with a score used for classification. Negative scores will be classified as ``stay'' and positive ones as ``abandon.'' The larger the absolute value of a score, the more likely its resulting class label. Since all continuous features were z-score normalized before classification, the actual input to classifiers are different than the values in the tree splits shown in Figure \ref{fig:tree}. The topmost splitting node corresponds to the best predictor in this tree. In this case, \textit{cml\_total\_dur} (appearing in the top row) bears more importance than \textit{cml\_idle\_time} and \textit{activated} (appearing in the middle row). 

\begin{figure}
\centering
\includegraphics[width=0.45\textwidth]{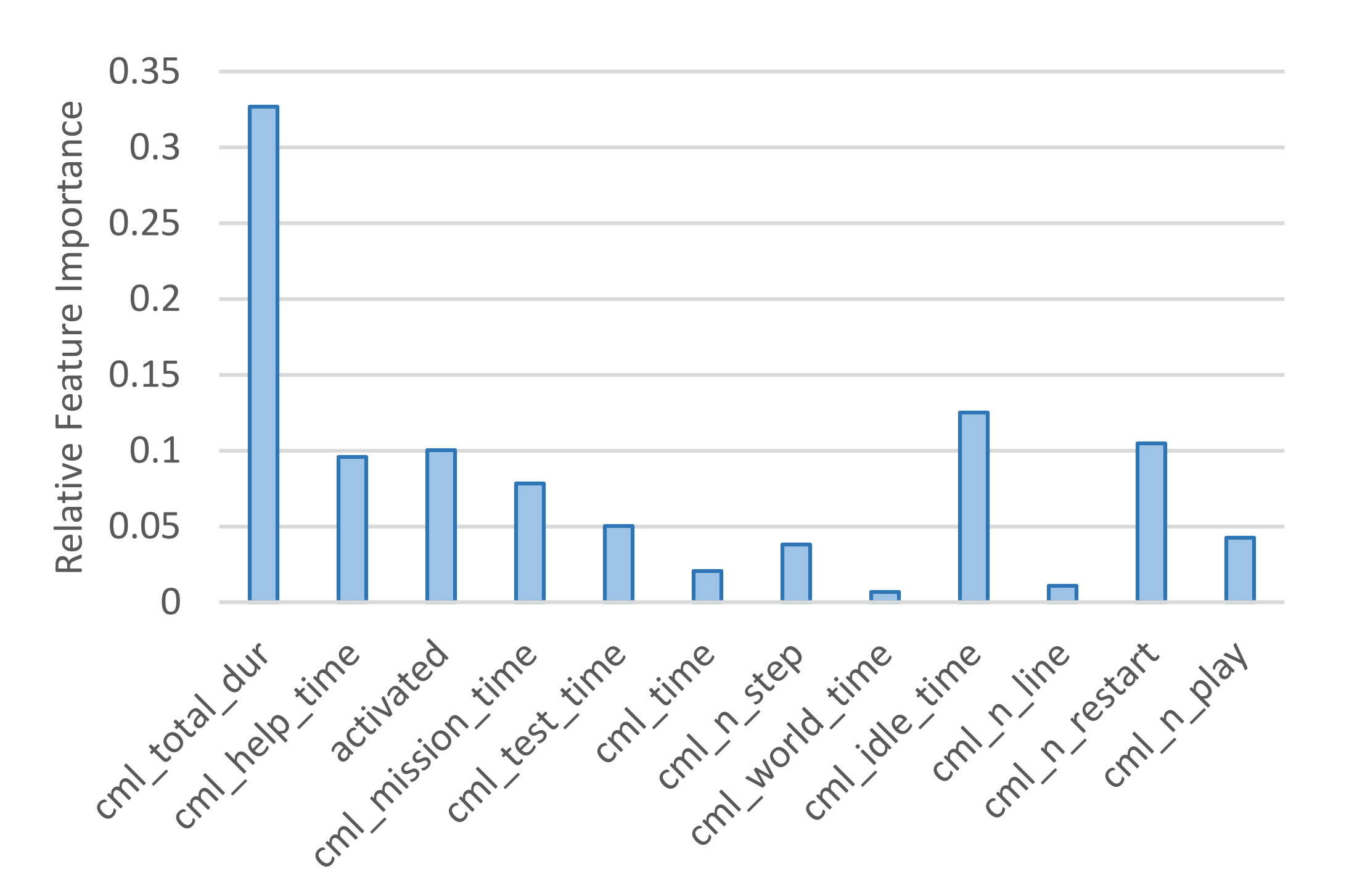}
 \caption{Level 5 classifier's relative importance of 12 features.}\label{fig:importance}
\end{figure}

To reveal a larger picture of feature importance assessed by all trees in a GBDT classifier, we repeated the process described in Section \ref{sec:method-importance} for each classifier and for each level, again obtaining relative importance measures. Figure \ref{fig:importance} shows the relative importance scores of Level 5, resulting in the following top five features: \textit{cml\_total\_dur} (0.327), \textit{cml\_idle\_time} (0.125), \textit{cml\_n\_restart} (0.104), \textit{activated} (0.100), and \textit{cml\_help\_time} (0.096). Based on these results, we know that these features are important in predicting abandonment (or staying) for this level, but it is difficult to interpret these results without knowing if these are positively or negatively related to abandonment. To address this issue, we examine these relationships below, in Section \ref{sec:impacts}.

Finally, examining the feature ranking across all five levels reveals there are two features with high relative importance across all levels: whether a learner had created an account (\textit{activated}) and cumulative non-idle time spent playing through the levels (\textit{cml\_total\_dur}). This suggests that the effort to fill out the registration form (which required an email address and a few demographic details) was a strong indicator of engagement and commitment. \textit{cml\_total\_dur} was also of high importance, suggesting that overall time actively interacting with the game was related to engagement. These results in the context of an online coding tutorial are consistent with findings from existing literature showing that indicators of commitment and/or effort have high predictive power in MOOCs or in-person scenarios \cite{greene_predictors_2015,jr_identifying_2005}.

\begin{table}[b]
  \caption{Odds ratio of each feature in each level.}
  \label{tab:OR}
  \centering
 \begin{tabular}{@{}llllll@{}}
\toprule
Feature                   & Lvl. 1  & Lvl. 2  & Lvl. 3  & Lvl. 4  & Lvl. 5 \\ \midrule
\rowcolor[HTML]{D3D3D3} 
cml\_help\_time           & 0.77(--) & 0.91(--) & 0.90(--) & 0.92(--) & 0.95(--) \\
\rowcolor[HTML]{D3D3D3} 
activated                 & 0.02(--) & 0.06(--) & 0.07(--) & 0.11(--) & 0.09(--) \\
\rowcolor[HTML]{A9A9A9} 
cml\_total\_dur           & 1.07(+) & 1.11(+) & 1.06(+) & 1.24(+) & 1.51(+) \\
\rowcolor[HTML]{A9A9A9} 
cml\_n\_step              & 1.18(+) & 1.03(+) & 1.25(+) & 1.10(+) & 1.44(+) \\
\rowcolor[HTML]{A9A9A9} 
cml\_idle\_time           & 1.21(+) & 1.38(+) & 1.33(+) & 1.11(+) & 1.30(+) \\
\rowcolor[HTML]{A9A9A9} 
cml\_n\_restart           & 1.05(+) & 1.28(+) & 1.11(+) & 1.14(+) & 1.27(+) \\
cml\_mission\_time        & 0.94(--) & 1.06(+) & 1.08(+) & 0.99(--) & 1.35(+) \\
cml\_test\_time           & 1.10(+) & 1.03(+) & 1.00(+) & 0.94(--) & 0.90(--) \\
cml\_code\_time           & 0.89(--) & 0.99(--) & 1.03(+) & 0.92(--) & 0.74(--) \\
cml\_world\_time 		  & 0.96(--) & 0.99(--) & 0.87(--) & 1.07(+) & 0.95(--) \\
cml\_n\_line              & 0.98(--) & 0.93(--) & 0.98(--) & 0.99(--) & 1.12(+) \\
cml\_n\_play              & 0.94(--) & 0.91(--) & 1.08(+) & 0.98(--) & 1.10(+) \\ \bottomrule
\end{tabular}
\end{table}

\subsection{Feature Impact} \label{sec:impacts}

To examine the direction of impact each feature has on abandonment, we ran a logistic regression for each level. Table \ref{tab:OR} shows the odds ratios (OR) of each feature in each level of logistic regression. OR $>$ 1 suggests that there is a positive association between the feature and abandonment, OR $<$ 1 suggests a negative association, and OR = 1 means the feature did not have an impact on abandonment. The further the absolute value of OR away from 1, the stronger the association. Features that consistently had a positive association with abandonment were \textit{cml\_total\_dur}, \textit{cml\_n\_step}, \textit{cml\_idle\_time}, and \textit{cml\_n\_restart}, as highlighted in dark gray. Conversely, features that consistently had a negative association with abandonment were \textit{cml\_help\_time} and \textit{activated}, as highlighted in light gray.
\section{Discussion}
This study is the first to examine factors that predict abandonment in an online coding tutorial using machine learned classifiers. Overall, our results suggest several things. First, classifiers using the features in Table ~\ref{tab:features} can do a much better job than baseline classifiers in identifying learners likely to abandon. Even with a small dataset of a few thousand learners, we can effectively separate the ones that are likely to abandon from the others. Second, these features do not provide as much predictive power as some of the social, identity, and motivational features more easily obtained in classroom settings, nor do they compete with dropout prediction models in MOOC settings, which have a wider set of features and interactions data with other users and instructors. Although our classifiers would likely to achieve better performance with more complex features through proper feature engineering, we have demonstrated that even simple features have measurable predictive value.

These discoveries have several implications for the design of online coding tutorials. First and foremost, designers can begin exploring how to use targeted abandonment prediction to encourage and engage learners online. Tutorials could provide encouraging messages at the beginning of a lesson; they could use avatars to convey encouraging words when learners at risk of abandoning encounter difficulty; they could even provide more targeted, context-relevant encouragement in response to learners' progress. Designers could also use abandonment prediction to predict the likelihood of success on the next level or lesson, adapting future levels to reduce difficulty, review challenging concepts, or present additional instruction before the learner proceeds to the next level. Of course, we have little evidence about whether such interventions would be effective; our feasibility assessment presented in this paper enables researchers to evaluate such interventions.

Our results also suggest several improvements to the classifiers we explored in this paper. For example, future work could explore the use of real-time predictions, using data not only from prior lessons in a tutorial, but more granular data about the current lesson a learner is engaged in. For example, real-time features might consider code testing, inactivity, and help seeking features, looking for signs of frustration, disengagement, and confusion. Real-time predictions would pose new interface design challenges as well, as intervening at arbitrary points in a lesson might pose some of the same challenges presented by intelligent assistants like ``Clippy,'' who seemed to interrupt at just the wrong times. Future work could also explore automatic hint generation based on the current progress presented by the a learner. 

There are several limitations in this work. First, because Gidget is an interactive game, we had access to features of engagement that many less interactive coding tutorials do not have. Other tutorials may need to enhance their interactivity to instrument the features that we utilized. Second, some of the features we used were dependent on each other, undermining our ability to assess feature importance. Additionally, we could not predict beyond Gidget's Level 6 due to a lack of sufficient data. There may be other interesting and different predictive possibilities for learners that persist to more advanced parts of coding tutorials, but we could not analyze these. We hope to do so in future work.

\section{Conclusion}
In this paper, we predicted abandonment of an online coding tutorial using machine learning classifiers, examined the importance of features that informed this prediction, and explored how these features associated with abandonment. Our results show that even with a small dataset and a few features related to engagement, classifiers can target encouragement to an average of 70\% of learners likely to abandon the next level, with the tradeoff of unnecessarily encouraging about 37\% of learners who will complete the next level. We found that these models gain their predictive power primarily from features related to \textit{commitment} (e.g., account activation) and \textit{effort} (non-idle time spent on all activities). For example, features that were consistently and positively associated with abandonment---\textit{cml\_total\_dur}, \textit{cml\_n\_step}, \textit{cml\_idle\_time}, and \textit{cml\_n\_restart}---likely characterized learners that spent a lot of time on the game, clicked the single execution button many times, had long periods of inactivity, and restarted/retried the level many times. Features that were consistent negatively associated with abandonment---\textit{cml\_help\_time} and \textit{activated}---likely characterized learners who were motivated enough to save their progress and seek further instruction from the game's help features. 
 
Ultimately, prediction efforts such as ours aim to provide some surrogate for the type of personalized, meaningful encouragement that teachers provide to their students. We are unlikely to ever have enough teachers online to provide encouragement to the tens of millions of people trying to learn to code from interactive tutorials. If we can provide even a fraction of this encouragement through automation, however, we might just nudge millions of people to the next lesson, further democratizing computing education.

\section*{Acknowledgment}
This work was supported in part by the National Science Foundation (NSF) under grants IIS-1657160, CNS-1240786, CNS-1240957, CNS-1339131, CCF-0952733, CCF-1339131, IIS-1314399, and IIS-1314384. Any opinions, findings, conclusions or recommendations are those of the authors and do not necessarily reflect the views of NSF.




%
\bibliographystyle{IEEEtran}
\bibliography{IEEEabrv,IEEEexample}

\begin{thebibliography}{10}
\providecommand{\url}[1]{#1}
\csname url@samestyle\endcsname
\providecommand{\newblock}{\relax}
\providecommand{\bibinfo}[2]{#2}
\providecommand{\BIBentrySTDinterwordspacing}{\spaceskip=0pt\relax}
\providecommand{\BIBentryALTinterwordstretchfactor}{4}
\providecommand{\BIBentryALTinterwordspacing}{\spaceskip=\fontdimen2\font plus
\BIBentryALTinterwordstretchfactor\fontdimen3\font minus
  \fontdimen4\font\relax}
\providecommand{\BIBforeignlanguage}[2]{{%
\expandafter\ifx\csname l@#1\endcsname\relax
\typeout{** WARNING: IEEEtran.bst: No hyphenation pattern has been}%
\typeout{** loaded for the language `#1'. Using the pattern for}%
\typeout{** the default language instead.}%
\else
\language=\csname l@#1\endcsname
\fi
#2}}
\providecommand{\BIBdecl}{\relax}
\BIBdecl

\bibitem{kizilcec_attrition_2015}
R.~F. Kizilcec and S.~Halawa, ``Attrition and achievement gaps in online
  learning,'' in \emph{{Conference} on {Learning}@{Scale}}.\hskip 1em plus
  0.5em minus 0.4em\relax ACM, 2015, pp. 57--66.

\bibitem{lee_-game_2013}
M.~J. Lee, A.~J. Ko, and I.~Kwan, ``In-game assessments increase novice
  programmers' engagement and level completion speed,'' in \emph{Conference on
  {International} Computing Education Research}.\hskip 1em plus 0.5em minus
  0.4em\relax ACM, 2013, pp. 153--160.

\bibitem{lee_personifying_2011}
M.~J. Lee and A.~J. Ko, ``Personifying programming tool feedback improves
  novice programmers' learning,'' in \emph{Workshop on International
  {Computing} Education Research}.\hskip 1em plus 0.5em minus 0.4em\relax ACM,
  2011, pp. 109--116.

\bibitem{gutl_attrition_2014}
C.~Gütl, R.~H. Rizzardini, V.~Chang, and M.~Morales, ``Attrition in {MOOC}:
  {Lessons} learned from drop-out students,'' in \emph{International {Workshop}
  on {Learning} {Technology} for {Education} in {Cloud}}.\hskip 1em plus 0.5em
  minus 0.4em\relax Springer, 2014, pp. 37--48.

\bibitem{hitz_praise_1988}
R.~Hitz and A.~Driscoll, ``Praise or encouragement? {New} insights into praise:
  {Implications} for early childhood teachers.'' \emph{Young Children}, 1988.

\bibitem{stake_critical_2006}
J.~E. Stake, ``The {Critical} {Mediating} {Role} of {Social} {Encouragement}
  for {Science} {Motivation} and {Confidence} {Among} {High} {School} {Girls}
  and {Boys},'' \emph{Journal of Applied Social Psychology}, vol.~36, no.~4,
  pp. 1017--1045, 2006.

\bibitem{tuckman_effect_1991}
B.~W. Tuckman and T.~L. Sexton, ``The effect of teacher encouragement on
  student self-efficacy and motivation for self-regulated performance,''
  \emph{Journal of Social Behavior and Personality}, vol.~6, no.~1, p. 137,
  1991.

\bibitem{brown_positive_2014}
L.~N. Brown and A.~M. Howard, ``The positive effects of verbal encouragement in
  mathematics education using a social robot,'' in \emph{Integrated {STEM}
  {Education} {Conference} ({ISEC}), 2014 {IEEE}}.\hskip 1em plus 0.5em minus
  0.4em\relax IEEE, 2014, pp. 1--5.

\bibitem{guzdial_teaching_2002}
M.~Guzdial and E.~Soloway, ``Teaching the {Nintendo} generation to program,''
  \emph{Communications of the ACM}, vol.~45, no.~4, pp. 17--21, 2002.

\bibitem{kinnunen_why_2006}
P.~Kinnunen and L.~Malmi, ``Why students drop out {CS}1 course?'' in
  \emph{Workshop on {Computing} Education Research}.\hskip 1em plus 0.5em minus
  0.4em\relax ACM, 2006, pp. 97--108.

\bibitem{bennedsen_failure_2007}
J.~Bennedsen and M.~E. Caspersen, ``Failure rates in introductory
  programming,'' \emph{ACM SIGCSE Bulletin}, vol.~39, no.~2, pp. 32--36, 2007.

\bibitem{watson_failure_2014}
C.~Watson and F.~W. Li, ``Failure rates in introductory programming
  revisited,'' in \emph{Conference on {Innovation} \& Technology in Computer
  Science Education}.\hskip 1em plus 0.5em minus 0.4em\relax ACM, 2014, pp.
  39--44.

\bibitem{ramalingam_self-efficacy_2004}
V.~Ramalingam, D.~LaBelle, and S.~Wiedenbeck, ``Self-efficacy and mental models
  in learning to program,'' in \emph{{ACM} {SIGCSE} {Bulletin}}, vol.~36.\hskip
  1em plus 0.5em minus 0.4em\relax ACM, 2004, pp. 171--175.

\bibitem{wilson_contributing_2001}
B.~C. Wilson and S.~Shrock, ``Contributing to success in an introductory
  computer science course: a study of twelve factors,'' in \emph{{ACM} {SIGCSE}
  {Bulletin}}, vol.~33.\hskip 1em plus 0.5em minus 0.4em\relax ACM, 2001, pp.
  184--188.

\bibitem{jr_identifying_2005}
P.~R.~V. Jr, ``Identifying predictors of success for an objects-first {CS}1,''
  \emph{Computer Science Education}, vol.~15, no.~3, pp. 223--243, Sep. 2005.

\bibitem{ramesh_modeling_2013}
A.~Ramesh, D.~Goldwasser, B.~Huang, H.~Daumé~III, and L.~Getoor, ``Modeling
  learner engagement in {MOOCs} using probabilistic soft logic,'' in
  \emph{{NIPS} {Workshop} on {Data} {Driven} {Education}}, vol.~21, 2013,
  p.~62.

\bibitem{taylor_likely_2014}
C.~Taylor, K.~Veeramachaneni, and U.-M. O'Reilly, ``Likely to stop?
  {Predicting} stopout in massive open online courses,'' \emph{arXiv preprint
  arXiv:1408.3382}, 2014.

\bibitem{balakrishnan_predicting_2013}
G.~Balakrishnan and D.~Coetzee, ``Predicting student retention in massive open
  online courses using hidden markov models,'' \emph{Electrical Engineering and
  Computer Sciences University of California at Berkeley}, 2013.

\bibitem{adamopoulos_what_2013}
P.~Adamopoulos, ``What makes a great {MOOC}? {An} interdisciplinary analysis of
  student retention in online courses,'' 2013.

\bibitem{hermans2017teaching}
F.~Hermans and E.~Aivaloglou, ``Teaching software engineering principles to
  k-12 students: a mooc on scratch,'' in \emph{Conference on Software
  Engineering: Software Engineering and Education Track}.\hskip 1em plus 0.5em
  minus 0.4em\relax IEEE Press, 2017, pp. 13--22.

\bibitem{yang_turn_2013}
D.~Yang, T.~Sinha, D.~Adamson, and C.~P. Rosé, ``Turn on, tune in, drop out:
  {Anticipating} student dropouts in massive open online courses,'' in
  \emph{{NIPS} {Data}-driven Education Workshop}, vol.~11, 2013, p.~14.

\bibitem{greene_predictors_2015}
J.~A. Greene, C.~A. Oswald, and J.~Pomerantz, ``Predictors of retention and
  achievement in a massive open online course,'' \emph{American Educational
  Research Journal}, p. 0002831215584621, 2015.

\bibitem{xing_temporal_2016}
W.~Xing, X.~Chen, J.~Stein, and M.~Marcinkowski, ``Temporal predication of
  dropouts in {MOOCs}: {Reaching} the low hanging fruit through stacking
  generalization,'' \emph{Computers in Human Behavior}, vol.~58, pp. 119--129,
  May 2016.

\bibitem{halawa_dropout_2014}
S.~Halawa, D.~Greene, and J.~Mitchell, ``Dropout prediction in {MOOCs} using
  learner activity features,'' \emph{Experiences and best practices in and
  around MOOCs}, vol.~7, 2014.

\bibitem{lee_comparing_2015-1}
M.~J. Lee and A.~J. Ko, ``Comparing the effectiveness of online learning
  approaches on {CS}1 learning outcomes,'' in \emph{{Conference} on
  {International} {Computing} {Education} {Research}}.\hskip 1em plus 0.5em
  minus 0.4em\relax ACM, 2015, pp. 237--246.

\bibitem{repenning2016retention}
A.~Repenning, A.~Basawapatna, D.~Assaf, C.~Maiello, and N.~Escherle,
  ``Retention of flow: Evaluating a computer science education week activity,''
  in \emph{Technical Symposium on Computing Science Education}.\hskip 1em plus
  0.5em minus 0.4em\relax ACM, 2016, pp. 633--638.

\bibitem{cocea_eliciting_2007}
M.~Cocea and S.~Weibelzahl, ``Eliciting motivation knowledge from log files
  towards motivation diagnosis for {Adaptive} {Systems},'' in
  \emph{International {Conference} on {User} {Modeling}}.\hskip 1em plus 0.5em
  minus 0.4em\relax Springer, 2007, pp. 197--206.

\bibitem{qu_detecting_2005}
L.~Qu and W.~L. Johnson, ``Detecting the learner's motivational states in an
  interactive learning environment,'' in \emph{Conference on {Artificial}
  {Intelligence} in {Education}: {Supporting} {Learning} through {Intelligent}
  and {Socially} {Informed} {Technology}}.\hskip 1em plus 0.5em minus
  0.4em\relax IOS Press, 2005, pp. 547--554.

\bibitem{6344507}
M.~J. Lee and A.~J. Ko, ``Investigating the role of purposeful goals on
  novices' engagement in a programming game,'' in \emph{2012 IEEE Symposium on
  Visual Languages and Human-Centric Computing (VL/HCC)}, Sept 2012, pp.
  163--166.

\bibitem{lee2014principles}
M.~J. Lee, F.~Bahmani, I.~Kwan, J.~LaFerte, P.~Charters, A.~Horvath, F.~Luor,
  J.~Cao, C.~Law, M.~Beswetherick \emph{et~al.}, ``Principles of a
  debugging-first puzzle game for computing education,'' in \emph{Visual
  Languages and Human-Centric Computing (VL/HCC)}.\hskip 1em plus 0.5em minus
  0.4em\relax IEEE, 2014, pp. 57--64.

\bibitem{jernigan_principled_2015}
W.~Jernigan, A.~Horvath, M.~Lee, M.~Burnett, T.~Cuilty, S.~Kuttal, A.~Peters,
  I.~Kwan, F.~Bahmani, and A.~Ko, ``A principled evaluation for a principled
  {Idea} {Garden},'' in \emph{Visual {Languages} and {Human}-{Centric}
  {Computing} ({VL}/{HCC})}.\hskip 1em plus 0.5em minus 0.4em\relax IEEE, 2015,
  pp. 235--243.

\bibitem{jernigan2017general}
W.~Jernigan, A.~Horvath, M.~Lee, M.~Burnett, T.~Cuilty, S.~Kuttal, A.~Peters,
  I.~Kwan, F.~Bahmani, A.~Ko \emph{et~al.}, ``General principles for a
  generalized idea gardenimage 1,'' \emph{Journal of Visual Languages \&
  Computing}, vol.~39, pp. 51--65, 2017.

\bibitem{lee2015teaching}
M.~J. Lee, ``Teaching and engaging with debugging puzzles,'' Ph.D.
  dissertation, 2015.

\bibitem{troyanskaya_missing_2001}
O.~Troyanskaya, M.~Cantor, G.~Sherlock, P.~Brown, T.~Hastie, R.~Tibshirani,
  D.~Botstein, and R.~B. Altman, ``Missing value estimation methods for {DNA}
  microarrays,'' \emph{Bioinformatics}, vol.~17, no.~6, pp. 520--525, 2001.

\bibitem{pedregosa_scikit-learn:_2011}
F.~Pedregosa, G.~Varoquaux, A.~Gramfort, V.~Michel, B.~Thirion, O.~Grisel,
  M.~Blondel, P.~Prettenhofer, R.~Weiss, V.~Dubourg, and {others},
  ``Scikit-learn: {Machine} learning in {Python},'' \emph{Journal of Machine
  Learning Research}, vol.~12, no. Oct, pp. 2825--2830, 2011.

\bibitem{chen_xgboost:_2016}
T.~Chen and C.~Guestrin, ``Xgboost: {A} scalable tree boosting system,'' in
  \emph{{Conference} on {Knowledge} {Discovery} and {Data} {Mining}}.\hskip 1em
  plus 0.5em minus 0.4em\relax ACM, 2016, pp. 785--794.

\bibitem{friedman_greedy_2001}
J.~H. Friedman, ``Greedy function approximation: a gradient boosting machine,''
  \emph{Annals of statistics}, pp. 1189--1232, 2001.

\bibitem{liang_machine_2016}
J.~Liang, C.~Li, and L.~Zheng, ``Machine learning application in {MOOCs}:
  {Dropout} prediction,'' in \emph{11th {International} {Conference} on
  {Computer} {Science} {Education} ({ICCSE})}, Aug. 2016, pp. 52--57.

\bibitem{thammasiri_critical_2014}
D.~Thammasiri, D.~Delen, P.~Meesad, and N.~Kasap, ``A critical assessment of
  imbalanced class distribution problem: {The} case of predicting freshmen
  student attrition,'' \emph{Expert Systems with Applications}, vol.~41, no.~2,
  pp. 321--330, Feb. 2014.

\bibitem{huang_telco_2015}
Y.~Huang, F.~Zhu, M.~Yuan, K.~Deng, Y.~Li, B.~Ni, W.~Dai, Q.~Yang, and J.~Zeng,
  ``Telco churn prediction with big data,'' in \emph{{SIGMOD} {International}
  {Conference} on {Management} of {Data}}.\hskip 1em plus 0.5em minus
  0.4em\relax ACM, 2015, pp. 607--618.

\bibitem{ho_random_1995}
T.~K. Ho, ``Random decision forests,'' in \emph{Document {Analysis} and
  {Recognition}, 1995., {Proceedings} of the {Third} {International}
  {Conference} on}, vol.~1.\hskip 1em plus 0.5em minus 0.4em\relax IEEE, 1995,
  pp. 278--282.

\bibitem{he_practical_2014}
X.~He, J.~Pan, O.~Jin, T.~Xu, B.~Liu, T.~Xu, Y.~Shi, A.~Atallah, R.~Herbrich,
  S.~Bowers, and {others}, ``Practical lessons from predicting clicks on ads at
  facebook,'' in \emph{{Workshop} on {Data} {Mining} for {Online}
  {Advertising}}.\hskip 1em plus 0.5em minus 0.4em\relax ACM, 2014, pp. 1--9.

\bibitem{szumilas_explaining_2010}
M.~Szumilas, ``Explaining {Odds} {Ratios},'' \emph{Journal of the Canadian
  Academy of Child and Adolescent Psychiatry}, vol.~19, no.~3, pp. 227--229,
  Aug. 2010.

\bibitem{lobo_auc:_2008}
J.~M. Lobo, A.~Jiménez-Valverde, and R.~Real, ``{AUC}: a misleading measure of
  the performance of predictive distribution models,'' \emph{Global ecology and
  Biogeography}, vol.~17, no.~2, pp. 145--151, 2008.

\end{thebibliography}


\end{document}